\documentclass[10pt,twocolumn,letterpaper]{article}

\usepackage[pagenumbers]{cvpr} 

\usepackage{graphicx}
\usepackage{amsmath}
\usepackage{amssymb}
\usepackage{booktabs}
\usepackage{comment}
\usepackage{csquotes}

\usepackage[table]{xcolor} 

\usepackage{times}
\usepackage{epsfig}
\usepackage{graphicx}
\usepackage{amsmath}
\usepackage{amssymb}
\usepackage{booktabs}

\usepackage{caption}
\usepackage{subcaption}

\usepackage{dblfloatfix}
\usepackage{tikz}
\usetikzlibrary{fit}
\usepackage{nicematrix}
\usepackage{comment}
\usepackage{color,soul}

\usepackage{flushend}

\usepackage[pagebackref=true,breaklinks=true,letterpaper=true,colorlinks,bookmarks=false, linkcolor=]{hyperref}

\usepackage[capitalize]{cleveref}
\crefname{section}{Sec.}{Secs.}
\Crefname{section}{Section}{Sections}
\Crefname{table}{Table}{Tables}
\crefname{table}{Tab.}{Tabs.}

\renewcommand{\mkbegdispquote}[2]{%
  \begin{minipage}[t]{0.1\textwidth}#2\end{minipage}%
  \begin{minipage}[t]{0.915\columnwidth}%
}
\renewcommand{\mkenddispquote}[2]{\end{minipage}}

\usepackage{tabularx}
\usepackage{makecell}

\def\cvprPaperID{14869} 
\def\confName{CVPR}
\def\confYear{2024}

\newcommand{\PLH}{{\mkern-2mu\times\mkern-2mu}}

\begin{document}
\title{High-Level Parallelism and Nested Features for Dynamic Inference Cost \\and Top-Down Attention}

\author{Andr\'e Kelm, Niels Hannemann\thanks{Niels Hannemann, Bruno Heberle, and Lucas Schmidt contributed equally to this work as part of their bachelor thesis.}, Bruno Heberle$^*$, Lucas Schmidt$^*$, Tim Rolff, \\ Christian Wilms, Ehsan Yaghoubi, Simone Frintrop\\
Universität Hamburg\\
Hamburg, Germany\\ 
{\ttfamily\small andre.kelm@, 0hannema@informatik., bruno.heberle@, lucas.schmidt-1@studium., tim.rolff@, } \\ {\ttfamily\small christian.wilms@, ehsan.yaghoubi@,
simone.frintrop@uni-hamburg.de}
}
\maketitle

\begin{abstract}
This paper introduces a novel network topology that seamlessly integrates dynamic inference cost 
with a top-down attention mechanism, addressing two significant gaps in traditional deep learning models.
Drawing inspiration from human perception, we
combine sequential processing of generic low-level features with parallelism and nesting of high-level features.
This design not only reflects a 
finding from recent neuroscience research regarding - spatially and contextually distinct neural activations - in human cortex,
but also introduces a novel ``cutout'' technique: the ability to selectively activate 
only network segments of task-relevant categories
to optimize inference cost and eliminate the need for re-training.
We believe this paves the way for future network designs that are lightweight and adaptable, making them suitable for a wide range of applications, from compact edge devices to large-scale clouds.
Our proposed topology also comes with a built-in top-down attention mechanism,
which allows processing to be directly influenced by either enhancing or inhibiting category-specific high-level features, drawing parallels 
to the selective attention mechanism observed in human cognition. 
Using targeted external signals, we experimentally enhanced predictions
across all tested models.
In terms of dynamic inference cost our methodology can achieve an exclusion of up to $73.48\,\%$ of parameters and $84.41\,\%$ fewer giga-multiply-accumulate (GMAC) operations, analysis against comparative baselines show an average reduction of $40\,\%$ in parameters and $8\,\%$ in GMACs across the cases we evaluated.
\end{abstract}

\section{Introduction}
\label{sec:intro}
One of the superior capabilities of the human brain is the ability to focus and accelerate processing when high-level knowledge is available. Not only does this save us energy, but it also increases our effectiveness and allows us to act in a very targeted manner.
This perceptual ability is still challenging to achieve with current deep learning (DL) methods.
If we search for a book on the bookshelf, the salt on the table, or our friend in the crowd, the human visual system is able to focus on target-relevant features and speed up processing considerably: Wolfe's measurements of human visual perception show that the reaction speed of such a guided visual search is 
usually significantly higher compared to an unguided search \cite{Wolfe2021}. 
This aspect is so general that it can also be observed across modalities.
A recent neuroscience paper by Marian \etal~\cite{crossmodalinteraction}, which inspired this work, argues that 
in the presence of audio signals such as a ``meow'', which already allow semantic inferences about a searched object, humans can speed up their visual search and more quickly perceive the object, in this case, the cat.
Marian \etal assume that there is a contextual connection between modalities and thus, for example, visual category-specific features for objects can be specifically activated. 
Even with recent DL methods,
such behavior cannot be easily reproduced \cite{10.3389/fnint.2020.00010}.
Widely used attention mechanisms, such as those used in transformer models, tend to be bottom-up, i.e.,~they focus on the given data \cite{attentionisalluneed}. This kind of attention is mainly statistically driven by many images or patches because these are needed for the mechanism to learn which features are more important than others for a particular decision. 
However, since we want to incorporate cues from the cognitive level (``meow'' $\rightarrow$ ``cat'') that include or exclude certain categories in processing, top-down attention is what we need.
So the question is: How should we design an artificial neural network that has such focus and accelerated processing capability?
\setlength{\fboxrule}{0.5px} 
\setlength{\fboxsep}{0pt}  
\begin{figure}
  \begin{subfigure}[b]{0.107\textwidth}
  \fbox{%
    \includegraphics[clip, trim=1.97cm 1.97cm 1.97cm 1.97cm, width=\textwidth]
    {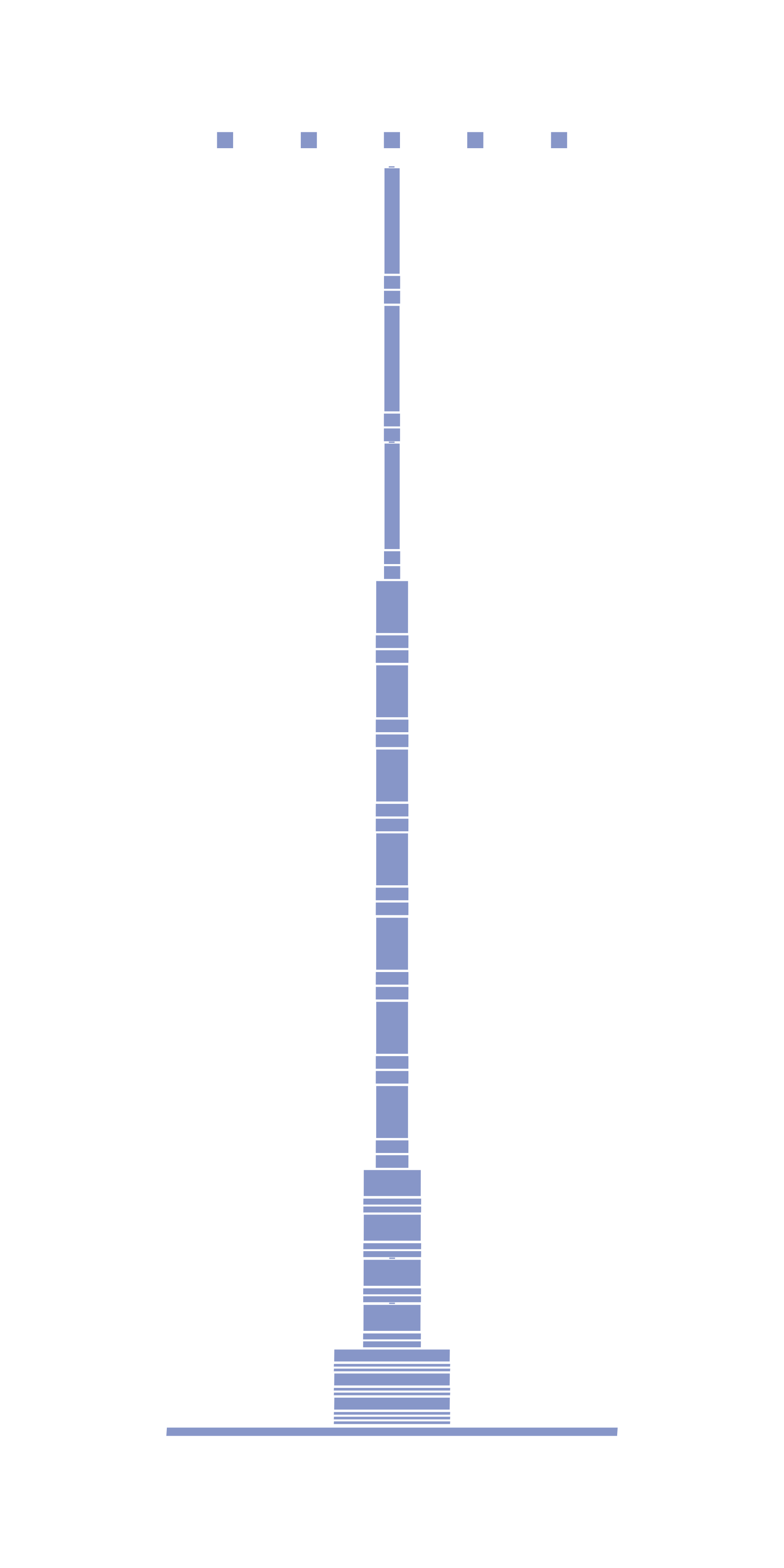}}
    \caption{Conventional}
    \label{fig:1a}
  \end{subfigure}
  \hfill
  \begin{subfigure}[b]{0.107\textwidth}
    \fbox{%
    \includegraphics[clip, trim=1.97cm 1.97cm 1.97cm 1.97cm, width=\textwidth]{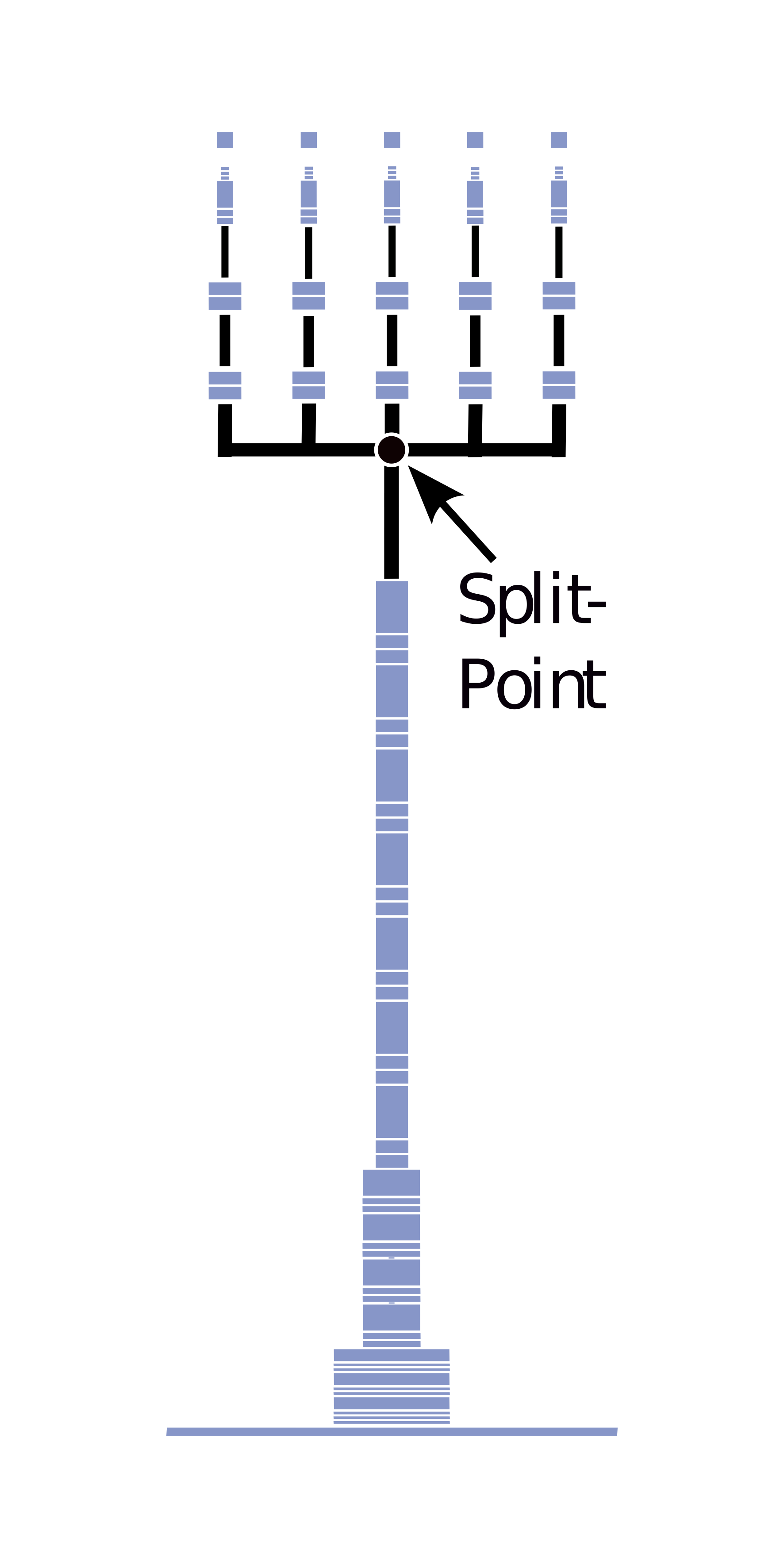}}
    \caption{SeqPar}
    \label{fig:2a}
  \end{subfigure}
  \hfill
  \begin{subfigure}[b]{0.107\textwidth}
    \fbox{%
    \includegraphics[clip, trim=1.97cm 1.97cm 1.97cm 1.97cm, width=\textwidth]{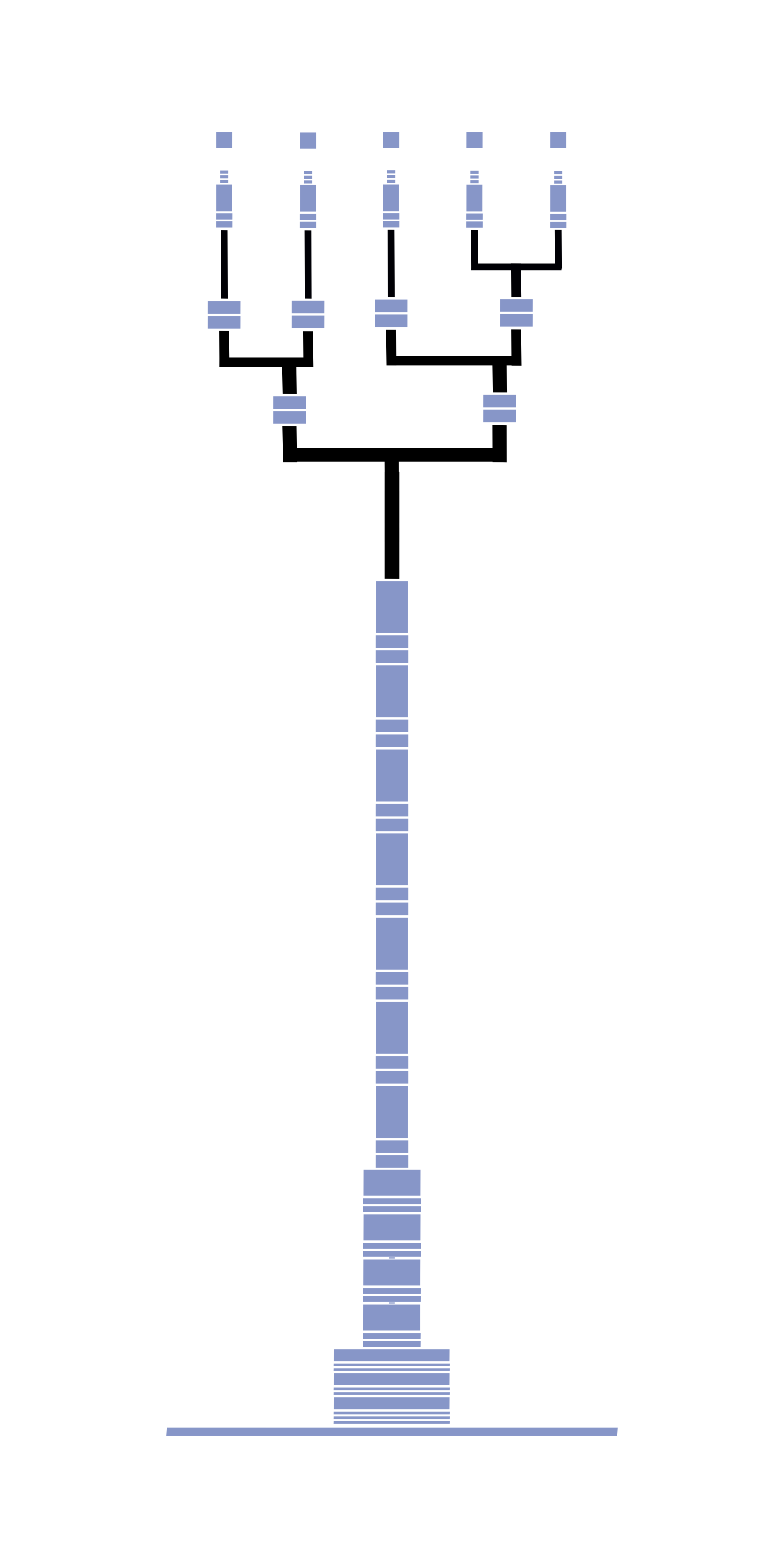}}
    \caption{Nested}
    \label{fig:3a}
  \end{subfigure}
    \hfill
  \begin{subfigure}[b]{0.107\textwidth}
    \fbox{%
    \includegraphics[clip, trim=1.97cm 1.97cm 1.97cm 1.97cm, width=\textwidth]{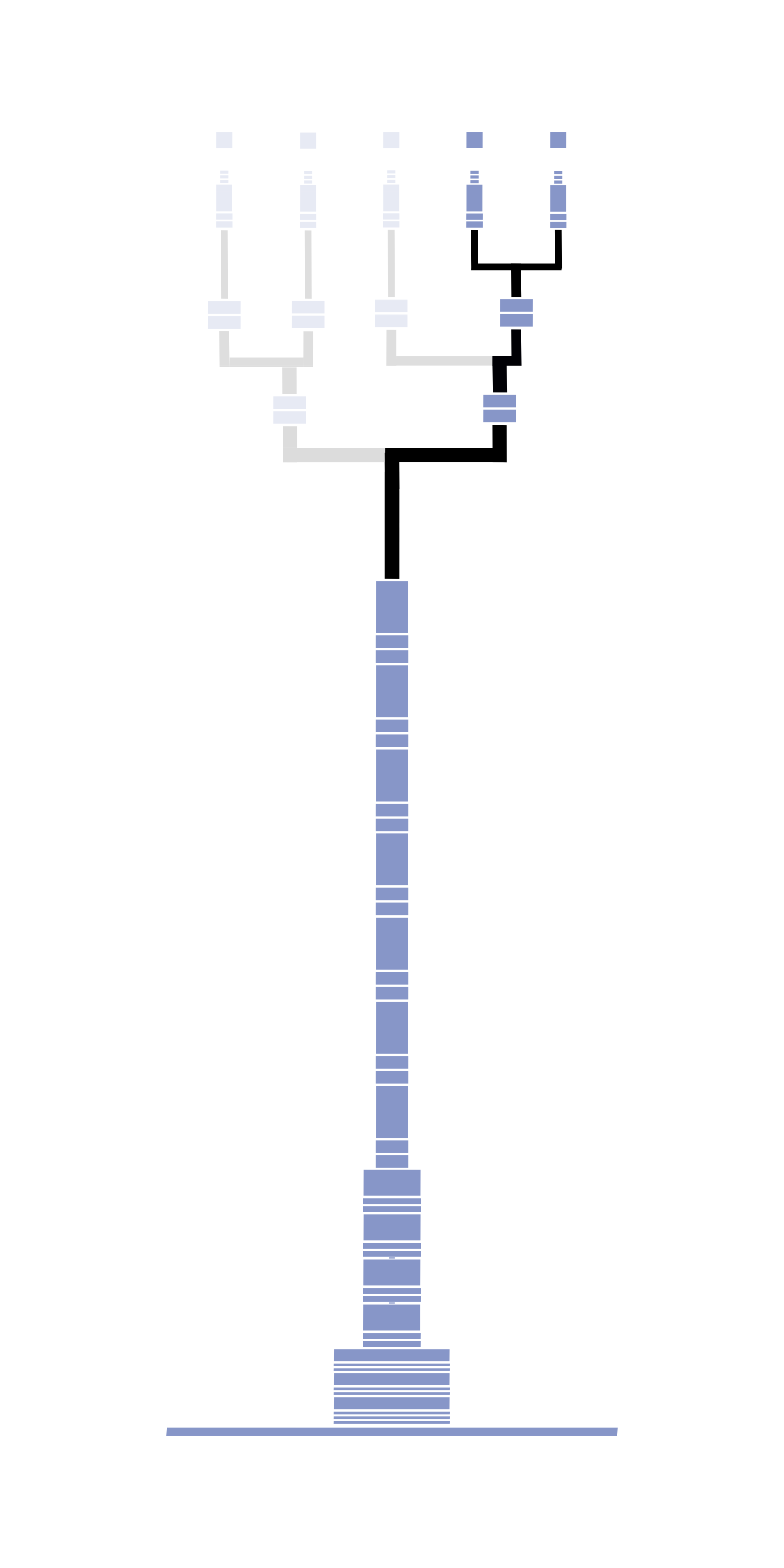}}
    \caption{Cutout}
    \label{fig:4a}
  \end{subfigure}
  \caption{Neural network topologies (blue: tensors, black: connection path). \subref{fig:1a} conventional deep network from bottom to top (the dots represent five categories); \subref{fig:2a} Our proposed SeqPar structure: each category has its own branch;
   \protect\subref{fig:3a} Our proposed compromise between a and b with a nested structure; \protect\subref{fig:4a} Our innovative cutout technique.}
  \label{Main figure}
\end{figure}

We assume that we should be able to exclude irrelevant features from the process to reduce processing time.
But DL methods are often described as black boxes \cite{aytekin}, so even the visualization of category-specific features is limited 
\cite{selvaraju2017gradcam}.
This property makes it challenging to use top-down cues, e.g.,~target or context information \cite{soubarna}, which could be useful to focus on relevant features and speed up processing.
Therefore, to access category-specific high-level features that are usually not directly addressable by top-down (context-derived) cues, we restructure, as shown in \autoref{Main figure}, the current convention (c.f.~\ref{fig:1a}) into the proposed \textbf{seq}uential-\textbf{par}allel (\textbf{SeqPar}) structure (c.f.~\ref{fig:2a}) and a nested approach (c.f.~\ref{fig:3a}).
Recent neuroscience studies suggest that layer 2 in the primary visual cortex responds in a context-specific way through spatially separated neurons \cite{Liu2022, doi:10.1073/pnas.1316808111, engramscomeofage}, 
similar to what we intend here:
Due to the parallelism of the layers towards the end of the network, features can be separated category-specifically \cite{featurespaceseparation}. 
They can now be amplified by simple scalar weighting, or processed by attention,
while low and mid-level features can still be shared by all categories.
Intuitively, this makes sense, since they are generic and useful for all categories.
But for high-level features it makes less sense because eyes are not useful for cars, for example. 

Our contributions are as follows:
\begin{itemize}

\item A novel SeqPar structure for category-specific cutouts, as shown in \autoref{fig:4a}, consisting of a fixed sequential component and a variable parallel component. These flexible cuts promote a more adaptive model with dynamic inference cost.
\item A built-in selective attention mechanism that allows the inference process of an end-to-end learned network to be actively influenced by external signals.
\item 
Nested networks in several variants, allowing the method to be applied to a larger number of categories.
\end{itemize} 

Overall, the SeqPar structure generates category-specific high-level features \cite{10.1007/978-3-030-93247-3_18} for another step towards interpretable AI, while providing the ability to bypass unneeded high-level features to reduce parameters, computational cost, and power consumption during inference.
This is important for mobile computing, industrial, drone, or robotic applications \cite{WANG20221, 10.1145/3486618, 8416402, 8463284}, and because the cost of inference and the cost of training for AI models used in industry is up to 9:1 \cite{DESISLAVOV2023100857}.

\section{Related Work}
\subsection{Image Classification and Dynamic Inference}
In the field of computer vision (CV), image classification is a key task, with ImageNet \cite{deng2009imagenet} validation traditionally serving as a benchmark \cite{CocA}. Powerful models developed for this task are then often applied to a variety of other areas, from semantic segmentation to object detection and video recognition \cite{videoswin}. 
This field is evolving rapidly, so methods such as transformers \cite{Zhang2021NestedHT}, large language models \cite{chatgpt3}, knowledge distillation \cite{dino}, wild and large datasets \cite{10.1007/978-3-030-01216-8_12}, unsupervised learning \cite{SIMCLR}, and matching loss functions \cite{symmentropy}, etc. are constantly advancing the state-of-the-art. Larger and deeper models often offer an accuracy advantage \cite{eva}, but this leads to increasing training and inference costs \cite{DESISLAVOV2023100857}. Notably, the industry tends to prioritize reducing inference costs over training costs by a factor of $9$ \cite{DESISLAVOV2023100857}, so a great variety of work has been done to dynamize inference costs \cite{dynneunet}. Dynamic routing, dynamic blocks and channels trained end-to-end are promising methods \cite{dynblock}.
However, the possibility of dynamizing the inference process by activating only target-relevant features depending on the task and context is not so often taken into account.
Such an implemented adaption could save parameters, energy, and processing time.
This is where our method comes in, so that high-level features can be generated category-specifically by parallelization and excluded when not needed.

\subsection{Bottom-up vs. Top-Down Attention}
Selective attention is a mechanism of human perception that enables us to focus on regions of potential interest \cite{davis2004,10.3389/fnint.2022.856207}. These can be objects, colors, locations, sounds, or other patterns. It helps us to deal with the complexity of the world and to quickly find objects of interest. Attention is driven by two types of cues: bottom-up and top-down cues \cite{theeuwes2010}. Bottom-up cues are salient patterns that automatically attract attention, such as a red flower on green grass. Top-down attention, on the other hand, focuses on regions which are behaviorally relevant and is driven by pre-knowledge, goals, or expectations. Searching for our key, or the well-known cocktail-party effect \cite{arons1992review} are examples of top-down attention.

Attention mechanisms have become very popular in DL \cite{Correia2022AttentionPA}.
However,
most of them cannot focus in a top-down manner like biological systems. They usually base their processing on the input data and do not consider pre-knowledge about the target or the scene. Early works before the DL era have amplified target-relevant features by excitation and inhibition of pre-computed features to realize top-down attention \cite{navalpakkam2005,frintrop2005}.
In deep networks, top-down attention is usually understood as a tracing of activations backwards through the network from class nodes to the feature maps, as in GradCAM \cite{selvaraju2017gradcam}, and is often used to localize class-specific features in feature maps. However, this requires one forward-pass of the input image through the network, before the backward tracing can be performed, and does not enable a quicker processing through the usage of pre-knowledge as in human perception. 

\subsection{Parallelism and Tree Networks}
Parallelism, as we introduce it in this work, is rarely used and, as far as we know, unique to this extent and in the context of dynamic inference or selective attention.
At first sight, it's reminiscent of ensemble learning, where multiple neural networks are trained side by side, e.g., to solve a task very precisely \cite{ensamble, ensamblearxive}, but it differs in the following point:
Ensembles often use multiple models, sometimes with specialized input, to train experts, whereas we train a single model with multiple branches and input from all categories,
characterized by a \mbox{SeqPar} structure. 
Thus, low-level and mid-level features can still be shared by all categories, while high-level features are separated in a category-specific way.

A work more related to ours than previously listed ones is the tree-like branching (TLB) network from Xue \etal~\cite{10.1007/978-3-030-93247-3_18} with a category-specific branching. 
They propose a tree structure based on the similarities of the categories and show improvements for image classification.
They do not draw parallels to brain inspiration or a relation to selective attention and therefore do not perform experiments in this direction, and although they show improved inference costs, they do so for the model as a whole and do not use the novelty of inference cost dynamization by cutouts introduced here. The architecture, nesting, and approach to nesting are also different.

\section{Proposed Method}
Current DL methods, unlike biological systems, cannot easily accelerate or focus their processing in scenarios where high-level knowledge from other modalities or insights are available \cite{crossmodalinteraction}.
They typically lack mechanisms for dynamically adjusting inference costs \cite{fullydyninf} or incorporating top-down attention \cite{soubarna}, which would intuitively require focusing on specific features.
Due to their ``black-box'' nature \cite{aytekin}, most models do not even reveal where category-specific features are located in the deep network.
To overcome these challenges,
we rethink the conventional, mostly sequential network topology and propose a SeqPar system that provides distinct and assignable layers for different categories.
Recent neuroscientific studies support our approach and show that the human brain also has context-specific areas \cite{Liu2022, doi:10.1073/pnas.1316808111}.

\subsection{High-Level Feature Parallelization}  
\label{ssec:NetArchi}
\begin{table}
\begin{center}
\resizebox{\columnwidth}{!}{%
\begin{NiceTabular}{|l|c|c|c|c|}
\hline
lay.\textbackslash mod. &  PHL$_\mathrm{small}$    & PHL$_\mathrm{big}$     & PHL$_\mathrm{lateSP}$ & ResNet50 \\
\hline\hline
input      &    \multicolumn{4}{c|}{$224 \times 224 \times 3$} \\
\hline
conv 1     &    \multicolumn{4}{c|}{$7 \times 7 \times 3 \times 64$, stride 2}  \\
pool 1     &    \multicolumn{4}{c|}{$3 \times 3$, max}  \\
\hline
output     &    \multicolumn{4}{c|}{$56 \times 56$}  \\
\hline
conv 2a    &    \multicolumn{4}{c|}{$1\times1\times64$}  \\
conv 2b    &    \multicolumn{4}{c|}{$3\times3\times64$}  \\
conv 2c    & \multicolumn{4}{c|}{$1\times1\times64\times64/256$}  \\
\hline
reps&\multicolumn{4}{c|}{$3$}       \\
\hline
output     &     \multicolumn{4}{c|}{$28 \times 28$}  \\
\hline
conv 3a    & \multicolumn{4}{c|}{$1\times1\times256\times128$}  \\
conv 3b    &  \multicolumn{4}{c|}{$3\times3\times128\times128$}  \\
conv 3c    &   \multicolumn{4}{c|}{$1\times1\times128\times128/512$}  \\
\hline
reps& \multicolumn{4}{c|}{$4$}    \\
\hline
output     &     \multicolumn{4}{c|}{$14 \times 14$}  \\
\hline
conv 4a    & $1\PLH1\PLH512\PLH128$ &\multicolumn{3}{c|}{$1\PLH1\PLH512\PLH256$} \\
conv 4b    & $3\PLH3\PLH128\PLH128$      & \multicolumn{3}{c|}{$3\PLH3\PLH256\PLH 256$} \\
conv 4c    & $1\PLH1\PLH128\PLH512$  & \multicolumn{3}{c|}{$1\PLH1\PLH256\PLH256/1024$} \\
\hline
reps &\multicolumn{2}{c|}{$1$} & \multicolumn{2}{c|}{$6$}       \\
\hline
par        &\multicolumn{2}{c|}{$k$}  &\multicolumn{2}{c|}{$1$}     \\
\hline
output     &     \multicolumn{4}{c|}{$7 \PLH 7$}  \\
\hline
conv 5a   & $1\PLH1\PLH512\PLH64$ &\multicolumn{2}{c|}{$1\PLH1\PLH1024\PLH128$}   & $1\PLH1\PLH1024\PLH512$  \\
conv 5b   & $3\PLH3\PLH64\PLH64$             & \multicolumn{2}{c|}{$3\PLH3\PLH128\PLH128$}  &$3\PLH3\PLH512\PLH512$ \\
conv 5c   & $1\PLH1\PLH64\PLH256$    &   \multicolumn{2}{c|}{$1\PLH1\PLH128\PLH512$}   & $1\PLH1\PLH256\PLH512/2048$ \\
\hline
reps   &\multicolumn{3}{c|}{$1$}       & $3$       \\
\hline
Par &\multicolumn{2}{c|}{$k$}               & $k$  & $1$     \\
\hline
output     &     \multicolumn{4}{c|}{$3 \times 3$}  \\
\hline
conv 6a   & $1\PLH1\PLH256\PLH32$ &\multicolumn{2}{c|}{$1\PLH1\PLH512\PLH64$} &\\
conv 6b   & \multicolumn{1}{c|}{$3\PLH3\PLH32\PLH32$}   & \multicolumn{2}{c|}{$3\PLH3\PLH64\PLH64$}        &  \\
conv 6c   &  \multicolumn{1}{c|}{$1\PLH1\PLH32\PLH128$}&  \multicolumn{2}{c|}{$1\PLH1\PLH64\PLH256$ }  &  \\
\hline
reps & \multicolumn{3}{c|}{$1$}    &\\
\hline
par &\multicolumn{2}{c|}{$k$}               & $k$  & $1$     \\
\hline
output     & \multicolumn{2}{c|}{$1 \times 1$}      &  \multicolumn{2}{c|}{$3 \times 3$}  \\
\hline

conv 7a    &\multicolumn{2}{c|}{} & $1\PLH1\PLH256\PLH32$ & \\
 conv 7b        & \multicolumn{2}{c|}{ }         & $3\PLH3\PLH32\PLH32$ &   \\
 conv 7c        &  \multicolumn{2}{c|}{ }& $1\PLH1\PLH32\PLH128$ & \\
\hline
reps &\multicolumn{2}{c|}{}      &\multicolumn{1}{c|}{$1$} &    \\
\hline
par &        \multicolumn{2}{c|}{} &         $k$           &    \\
\hline
output        &  \multicolumn{3}{c|}{$1 \times 1$}  & \multicolumn{1}{c|}{$3 \times 3$}   \\
\hline
pool\&flat &     \multicolumn{4}{c|}{$3 \times 3$ average pooling $+$ flatten layer}   \\
\hline
FC   &   $128$      &  $256$    &  $128$        &  $2048$  \\
\hline
output     &   \multicolumn{4}{c|}{$1 \times 1 \times 1 \times k$}   \\
\hline
\CodeAfter
\tikz \node [draw=red, fit = (20-2) (20-2)] { } ;
\tikz \node [draw=red, fit = (26-4) (26-4)] { } ;
\tikz \node [draw=green, fit = (26-2) (26-2)] { } ;
\tikz \node [draw=green, fit = (32-4) (32-4)] { } ;
\tikz \node [draw=blue, fit = (38-4) (38-4)] { } ;
\end{NiceTabular}
}
\end{center}
\caption{Detailed model architectures. Left: three variants of our PHL method; right: a ResNet50 of conventional type. The red boxes indicate the location of the split-point which introduces the main change of the network topology: PHLs get as many parallel branches as the categories $k$ to be recognized; ResNets have only one.
On the other hand, PHL has fewer Seq repetitions (abbreviated as reps) of conv blocks - resulting in a shallower net.
The green box is the insertion point for a second hierarchy level for NHL$_\mathrm{wordnet}$ and the blue one marks a third hierarchy for NHL$_\mathrm{gpt}$.}
\label{tab:bigtab}
\end{table}
\begin{figure}[t!]
    \centering
    \includegraphics[height=0.1969\textwidth, clip, trim=0.225cm 0.225cm 0.225cm 0.225cm]{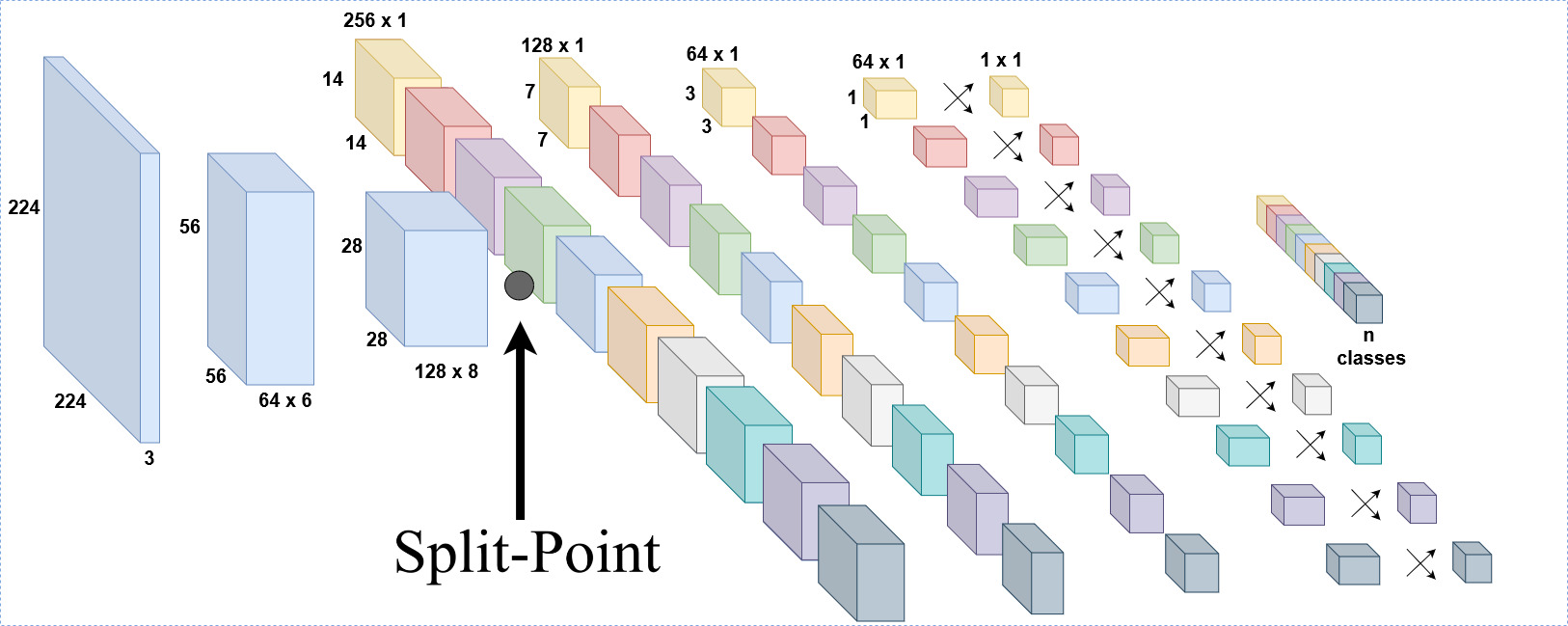}\label{fig:f2}
    \caption{Our PHL architecture starts sequentially, but becomes parallel after a split-point to generate category-specific features that are spatially separated from each other. The different colors represent the categories and their branches. The resolution and the dimension of the feature channel, multiplied by the number of layers have exemplary values for a better understanding.}
\label{fig:f2}
\end{figure}
We are exploring a new network topology, whose principle can be applied to modern models. Therefore, we have based our model on a widely known standard, the ResNet50. \autoref{fig:f2} shows our proposed 
\textbf{P}arallelize features at a \textbf{H}igh-\textbf{L}evel network (\textbf{PHL}).
We define the PHL by several blocks containing several layers. It can be divided into two parts: the first layers, which are classically sequential, and the deeper layers, which have the novel parallel information flow, all including ResNet skip connections. In \autoref{tab:bigtab} we have listed three variants.
They differ by having fewer (small) or more (big) channels per parallel branch or by a later split-point (lateSP). 
The hypothesis behind splitting the architecture into multiple branches while simultaneously processing the initial input through a sequential network is that lower-level features are generic and useful for almost any category, while higher-level features are more specific, e.g., eye features are not required for a car and should therefore be treated in parallel. We define this connection as the split-point that connects the two parts of the network.
Intuitively, a split-point could benefit from a larger number of parameters as it has to handle $1:k$ connections, so we use a bottleneck structure \cite{ResNet}.

\paragraph{Split-Point:}
\begin{figure}[t!]
    \centering
    \includegraphics[width=0.4725\textwidth]{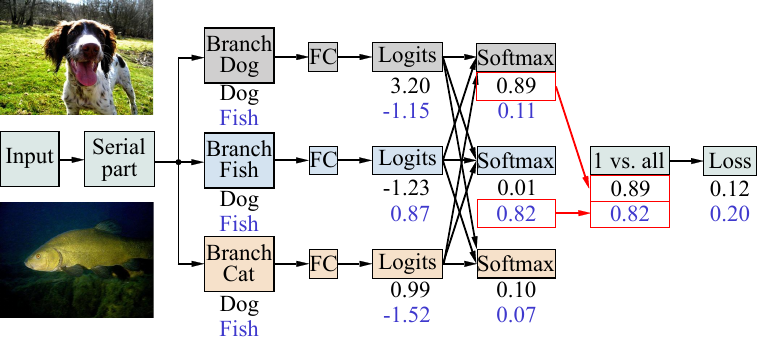}
    \caption{Three-branched PHL network for dog, fish and cat categories with some example values (black: input dog, blue: input fish). The red boxes indicate the selection for 1~vs.~all. Images from Imagenette \cite{imagenette}.}
\label{fig:CELoss}
\end{figure}
Formally, we define the first part of our architecture to consist of shared network layers up to the split-point through the function: $\phi(x, \theta_0)$, taking the input image $x$ and network parameters $\theta_0$ as input. As we provide a branch for each of the target categories $k$, we define the number of final branches to be the same as our $k$ target categories. 
Hence, we use $k$ branch networks that we define through $\psi(\bar{x}, \theta_i)$ with $1\leq i\leq k$, which
take the output features of $\phi$ as input for $\bar{x}$. As all branch networks have the same design, we only exchange the network parameters $\theta_i$ for the classification of the $i^\text{th}$ class. Using the above definitions allows us to define the output of the $i^\text{th}$ branch network, which is responsible for class $i$, as:
\begin{align}
    f_i(x) = \psi(\phi(x, \theta_0), \theta_i) = y_i.
\end{align}

\paragraph{Cross-Entropy Loss:}
For training of the network, we retain all initial target categories and estimate the class probability over all categories for the likelihood of an input $x$ belonging to a class 
and utilized a cross-entropy as our loss function. Note also that this is the special case of Eq.~\ref{eq: Softmax} where we utilized all categories instead of a subset.
The result of the likelihood estimation is included in a 1~vs.~all classification and the cross entropy loss, see details in \autoref{fig:CELoss}.
Optimizing the loss through backpropagation results in a high-value output for an image that belongs to the branch and a low-value output otherwise. Due to the softmax function, all paths are learned at once. \autoref{fig:CELoss} illustrates this central point of our approach.

\subsection{Nested Topology}
\label{ssec:Nested}
By pursuing the idea of category-dependent feature similarity and feature sharing at different hierarchical levels between image categories, we assume that a more optimal 
network 
topology can be constructed by nesting branches depending on the categories and their superclasses. We hypothesize that there are higher-level features that should be shared, such as wheels for buses and cars, and propose the \textbf{N}est features at a \textbf{H}igh-\textbf{L}evel (\textbf{NHL}) network.
\autoref{fig:3a} shows a schematic of such an architecture.
To get an additional level of hierarchy, we use the superclasses provided by imagenet-superclass \cite{s-kumano} and Wordnet \cite{wordnet} for ImageNet categories \cite{deng2009imagenet}.
Using these, we define this extended architecture by introducing another split-point, indicated by the green box for PHL-small in \autoref{tab:bigtab}. Here, multiple branches~$i$ share the features of a superclass~$j$ through an additional sub-network $\pi(\hat{x}, \hat\theta_j)$ with network parameters $\hat\theta$ and again taking the output of $\phi$ as input for $\hat{x}$ and passing its output to $\bar{x}$ of $\psi$. 
Hence, a total number of superclasses $s$ results in an equal number of branches $1\leq j\leq s$ in the lower hierarchical layer. 
Note that this does not change the total number of terminal branches $k$ in the upper hierarchical layer, since they are still equal to the number of categories in the dataset. This allows us to redefine the output of the hierarchical branching network:
\begin{align}
    \hat{f}_i(x) = \psi(\pi(\phi(x, \theta_0), \hat{\theta}_j), \theta_i) = y_i.
\end{align}
To train the network, we use the aforementioned cross-entropy loss and our redefined softmax.
In developing more nesting variants with three or four split-points in a fast and easy way, we consider several suggestions from OpenAI's ChatGPT$4$ \cite{OpenAI_ChatGPT}.


\paragraph{NHL$_\mathrm{gpt}$:}
In the case of just two split-points, there is a high number of possible branch nestings, exceeding $B_{100} \approx 10^{115}$ for 100 and $B_{1000} \approx 10^{1927}$ for 1000 categories, with $B_n$ being the Bell function \cite{tanny_1975}. Yet, for a network with just one split-point this number is equal to one, as it does not allow rearranging as the number of categories is fixed. Hence, for a three hierarchy levels deep network N3HL$_\mathrm{gpt}$, we employed a hybrid semi-automatic approach using ChatGPT. Here, we leveraged ChatGPT whenever the classification was ambiguous.

\subsection{Cutouts}
\label{ssec:cutouts}
\begin{figure}[h!]
    \centering
    \includegraphics[clip, trim=12.7mm 10.7mm 10.7mm 10.7mm, width=0.475\textwidth]{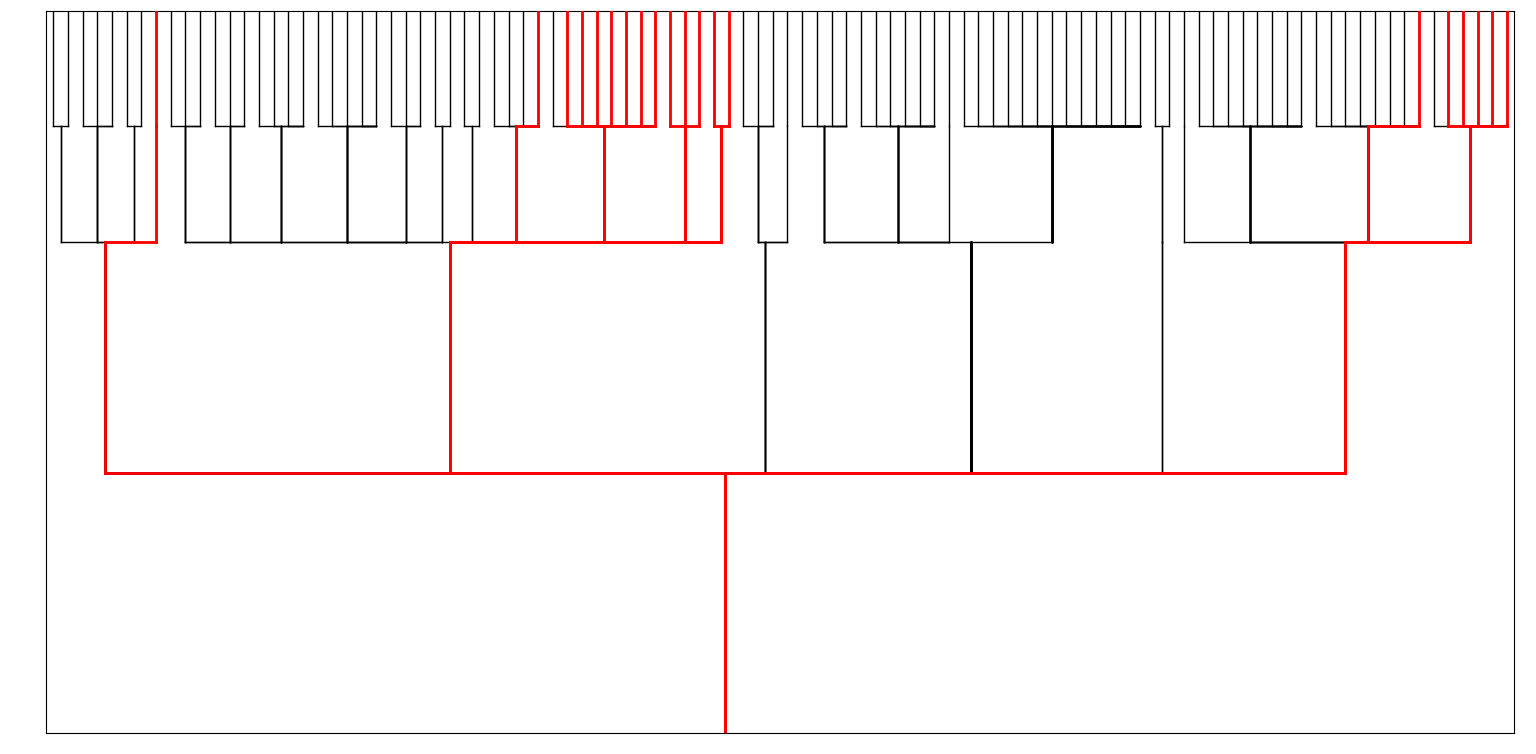}
    \caption{Cutout from a N$3$HL network with $100$ 
categories. The red pathways highlight the activated paths for the 
selected $20$ categories from ImageNet100 \cite{imagenette}, 
demonstrating the ``cutout'' technique (processing from bottom to top).}
\label{fig:cutout}
\end{figure}

Conventional architectures usually do not have the ability to omit category-specific high-level features that are not needed.
Often it is not even clear which parameter contributes to which category.
However, if only a subset of categories is needed for a specific task - whether by changed conditions or due to prior semantic knowledge - an option to exclude category-specific features would be advantageous.
Through the above made definition of a split-point, we can define cutouts that allow the network to adapt to simpler tasks in the case that prior knowledge is available. This enables the formation of selected category-specific paths that are shorter and faster than if a standard neural network has to be used (c.f.~\ref{fig:cutout}). 
Here, we extract $n\leq k$ selected paths using a-priori knowledge of potential $n$ new target categories $C=\left\{i_0, \cdots, i_n\ \middle|\ 1\leq i_j \leq k\right\}$, while simultaneously keeping the branch fully functional. This allows us to define the probability of a sample belonging to that category $i\in C$ through:
\begin{align}
    p_i(x) = \frac{e^{-f_i(x)}}{\sum_{j\in C} e^{-f_j(x)}}.
    \label{eq: Softmax}
\end{align}
The ability to adapt to the lower complexity of a task with less computational effort, but also to increase it when needed, is reminiscent of efficient and effective biological systems and their ability to focus.

\section{Experiments}
We evaluate five of our proposed topology variants PHL (see Sec.~\ref{ssec:NetArchi}), N2HL, N3HL and N4HL (see Sec.~\ref{ssec:Nested}) by implementing them using ResNet50 and the N3HL152 by using ResNet152 and compare them with their conventional counterpart.
We include ResNet18 and -200 to measure some lower and upper performance limits. 
Our unique top-down attention mechanism (see Sec.~\ref{ssec:selattinbuilt}) is tested for PHL and NHL, highlighted in light gray in the respective tables and underlined with TD attention. 
It is worth mentioning, that these are not separate models, as they are subject to
amplification of targeted feature maps by  external signals, this technique is applicable to all PHL or NHL networks.

\subsection{Training and Datasets}
\label{ssec:TrainData}
We have implemented all our models in the Timm library \cite{rw2019timm} and used Rand Augment \cite{randaugment}.
We used three different levels of classification complexity and training setups for studying PHL and NHL:
\begin{itemize}
    \item $10$ categories with $1000$ epochs, batch size $32$
    \item $100$ categories with $300$ epochs, batch size $56$, maxed out $24,564$~MebiBytes (MiB), mixed precision training
    \item $1000$ categories with $300$ epochs and batch size $24$, and another run with batch size $46$, maxed out $24,564$ and $49,140$~MiB, both trained through mixed precision training
\end{itemize}
As datasets, we chose subsets of ImageNet \cite{deng2009imagenet}.
Imagenette and Imagewoof \cite{imagenette} both contain $10$ categories, with the interesting property that Imagenette contains categories, which differ considerably, while Imagewoofs categories are quite similar (breeds of dogs). 
Testing these two datasets helps to understand how the parallel branches handle similar and non-similar categories. 
ImageNet100 \cite{deng2009imagenet, imagenet100_kaggle} has some similar and some differing categories, but shows how the model handles an upscaling by a factor of $10$.
The datasets used here have similar category and image numbers, and therefore a similar significance to Cifar10 and -100. These are not used because their lower image resolution would introduce an unwanted new factor into the evaluation \cite{cifar10, cifar100}.
ImageNet mini \cite{figotin_2020} is a subset of ImageNet1k and has only a few images per category, about $20$-$50$, but offers a high classification complexity with $1000$ categories. 
We evaluate the top-1 accuracy and omit the top-5, as this does not carry much meaning for the datasets with few categories. 
\subsection{Dynamic Inference Cost}
\label{ssec:cutoutseval}
\begin{table}[h!]
\begin{center}
\resizebox{\columnwidth}{!}{%
\begin{tabular}{|l||c|c||c|c|c|}
\hline
dataset & \multicolumn{4}{c|} {Imagenette} \\
\hline
model & \multicolumn{2}{c||} {ResNet50} & \multicolumn{2}{c|} {PHL$_\mathrm{lateSP}$  }\\
\hline
GMACs & \multicolumn{2}{c||} {$4.13$}  & \multicolumn{2}{c|} {$\textbf{3.99}$}\\
\hline
GMACs cutout& \multicolumn{2}{c||} {NA}  & \multicolumn{2}{c|} {$\textbf{3.66}$}\\
\hline
GMACs. reduction & \multicolumn{2}{c||} {$0\: \%$} & \multicolumn{2}{c|} {$\textbf{-8.40 \%}$}\\
\hline
parameter & $23,518,277$ & $23,528,522$ & \multicolumn{2}{c|} {$\textbf{19,986,506}$}\\
\hline
param. cutout & \multicolumn{2}{c||} {NA} & \multicolumn{2}{c|} {$\textbf{14,264,901}$}\\
\hline
param. reduction & \multicolumn{2}{c||} {$0\: \%$} & \multicolumn{2}{c|} {$\textbf{-28.63 \%}$}\\
\hline
train categories & upper 5  & all 10  & \multicolumn{2}{c|} {all 10} \\
val categories   & upper 5 &  upper 5 & \multicolumn{2}{c|} {upper 5}\\
\hline
top-1 accuracy& $97.16$ & $\bold{97.83}$ & \multicolumn{2}{c|} {$97.62$}\\
\hline
train categories & lower 5  & all 10  & \multicolumn{2}{c|} {all 10 }\\
val categories & lower 5  & lower 5  & \multicolumn{2}{c|} {lower 5} \\
\hline
top-1 accuracy& $96.79$ & $\bold{98.04}$ & \multicolumn{2}{c|} {$\bold{98.04}$}\\
\hline
\hline
dataset & \multicolumn{4}{c|} {Imagewoof} \\
\hline
model & \multicolumn{2}{c||} {ResNet50} & \multicolumn{2}{c|} {N$3$HL$_\mathrm{gpt}$  }\\
\hline
GMACs & \multicolumn{2}{c||} {$4.13$}  & \multicolumn{2}{c|} {$\textbf{3.91}$}\\
\hline
GMACs cutout& \multicolumn{2}{c||} {NA}  & \multicolumn{2}{c|} {$\textbf{3.73}$}\\
\hline
GMACs reduction& \multicolumn{2}{c||} {$0\: \%$}  & \multicolumn{2}{c|} {$\textbf{-4.90 \%}$}\\
\hline
parameter & $23,518,277$ & $23,528,522$ & \multicolumn{2}{c|} {$\textbf{11,929,930}$}\\
\hline
param. cutout & \multicolumn{2}{c||} {NA} & \multicolumn{2}{c|} {$\textbf{10,781,253}$}\\
\hline
param. reduction & \multicolumn{2}{c||} {$0\: \%$} & \multicolumn{2}{c|} {$\textbf{-9.63 \%}$}\\
\hline
train categories &  upper 5 & all 10 & \multicolumn{2}{c|} {all 10} \\
val categories&  upper 5  & upper 5  &  \multicolumn{2}{c|} {upper 5}  \\
\hline
top-1 accuracy& $91.18$ & $91.88$ & \multicolumn{2}{c|} {$\bold{92.69}$}\\
\hline
train categories&  lower 5  & all 10 & \multicolumn{2}{c|} {all 10}\\
val categories&  lower 5 & lower 5 &  \multicolumn{2}{c|} {lower 5}\\
\hline
top-1 accuracy& $\bold{96.23}$ & $94.30$ & \multicolumn{2}{c|} {$95.89$}\\
\hline
\hline
dataset & \multicolumn{4}{c|} {ImageNet100} \\
\hline
model & \multicolumn{2}{c||} {ResNet50} & \multicolumn{2}{c|} {N$3$HL$_\mathrm{gpt}$ }\\
\hline
GMACs & \multicolumn{2}{c||} {$\textbf{4.13}$}  & \multicolumn{2}{c|} {$5.86$}\\
\hline
GMACs cutout& \multicolumn{2}{c||} {NA}  & \multicolumn{2}{c|} {$4.22$}\\
\hline
GMACs. reduction & \multicolumn{2}{c||} {$0\: \%$} & \multicolumn{2}{c|} {$\textbf{-27.99 \%}$}\\
\hline
parameter & $\textbf{23,549,012}$ &$\textbf{23,712,932}$ & \multicolumn{2}{c|} {$24,950,436$}\\
\hline
param. cutout & \multicolumn{2}{c||} {NA} & \multicolumn{2}{c|} {$\textbf{13,840,469}$}\\
\hline
param. reduction & \multicolumn{2}{c||} {$0\: \%$} & \multicolumn{2}{c|} {$\textbf{-44.53 \%}$}\\
\hline
train categories & lower 20  & all 100 & \multicolumn{2}{c|} {all $100$}\\
val categories & lower 20 & lower 20 & \multicolumn{2}{c|} {lower $20$}\\
\hline
top-1 accuracy& $ 88.50 $ & $90.60$ & \multicolumn{2}{c|} { $\textbf{93.40}$}\\
\hline
\rowcolor{lightgray}
top-1$_\mathrm{TD\:attention}$& NA & NA & \multicolumn{2}{c|} { $\textbf{94.20}$}\\
\hline
dataset & \multicolumn{4}{c|} { ImageNet1k} \\
\hline
model & \multicolumn{2}{c||} {ConvNeXtV2} & \multicolumn{2}{c|} {ConvNeXtV2$_\mathrm{NHL}$ }\\
\hline
GMACs & \multicolumn{2}{c||} {$4.56$}  & \multicolumn{2}{c|} {$\textbf{14.55}$}\\
\hline
GMACs cutout& \multicolumn{2}{c||} {NA}  & \multicolumn{2}{c|} {$\textbf{3.87}$}\\
\hline
GMACs. reduction & \multicolumn{2}{c||} {$0\: \%$} & \multicolumn{2}{c|} {$\textbf{-73.40 \%}$}\\
\hline
parameter & $\textbf{27,801,029}$ &$\textbf{28,566,184}$ & \multicolumn{2}{c|} {$119,879,336$}\\
\hline
param. cutout & \multicolumn{2}{c||} {NA} & \multicolumn{2}{c|} {$\textbf{13,488,133}$}\\
\hline
param. reduction & \multicolumn{2}{c||} {$0\: \%$} & \multicolumn{2}{c|} {$\textbf{-88.74 \%}$}\\
\hline
train categories & meow $5$  & all $1000$ & \multicolumn{2}{c|} {all $1000$}\\
val categories & meow $5$ & meow $5$ & \multicolumn{2}{c|} {meow $5$}\\
\hline
top-1 accuracy& $\textbf{80.00}$ & $79.60$ & \multicolumn{2}{c|} { $\textbf{80.00}$}\\
\hline
\end{tabular}
}
\end{center}
\caption{Dynamic inference costs of six cutouts compared to ResNet50. For Imagenette and Imagewoof, the analysis is divided into two subsets: upper (first x categories) and lower (remaining y categories). ImageNet100 dataset is evaluated with a 20-category cutout, while ImageNet mini focuses on a cutout specific to cat.
\label{tab:cutouts}
}
\end{table}

\begin{table*}[t]
\centering
\resizebox{2.05\columnwidth}{!}{
\begin{tabular}{|l| ccc| cc| ccc|}
\multicolumn{1}{c}{\thead{Dataset}} & \multicolumn{1}{c}{\thead{}} & \multicolumn{1}{c}{\thead{Imagenette}} & \multicolumn{1}{c}{\thead{Imagewoof}} & \multicolumn{1}{c}{} & \multicolumn{1}{c}{\thead{ImageNet100}} & \multicolumn{1}{c}{} & \multicolumn{1}{c}{\thead{ImageNet mini}} & \multicolumn{1}{c}{\thead{ImageNet1k}} \\
\cmidrule(lr){1-1} \cmidrule(lr){2-4} \cmidrule(lr){5-6} \cmidrule(lr){7-9}

\thead{Metric} & no. parameter & top-1 acc. & top-1 acc. & no. parameter & top-1 acc. & no. parameter & top-1 acc. & top-1 acc. 
\\
\hline
\multicolumn{4}{l}{\emph{   }}\\
\hline
\hspace{0.2cm} PHL$_\mathrm{small}$ \textbf{(base 1)} & $9,611,338$ &               $95.71$ & $89.97$ & $83,109,028$ &           $82.52$ & $818,213,928$  & $21.10$ & -\\
\rowcolor{lightgray}
\hspace{0.2cm} PHL$_\mathrm{TD\:attention}$ & $9,611,338$ &  $ \textbf{96.38} $ & $ \textbf{90.54} $ & $83,109,028$ & $ \textbf{82.56} $ & $818,213,928$  & $ \textbf{21.34} $ & - \\
\hline
\multicolumn{4}{l}{\emph{   }}\\
\hline
\hspace{0.1cm} ResNet50 \textbf{(base 2)} & $23,528,522$ &                   $96.15$ & $90.30$ & $23,712,932$ & $85.34$ & $25,557,032$ & $\textbf{33.38}$  & - \\
\hspace{0.1cm} ResNet18 & $11,181,642$ &                   $95.52$ & $89.97$ & $11,227,812$ &            $81.76$ & $11,689,512$ & $28.52$ & - \\
\hspace{0.1cm} N$2$HL$_\mathrm{wordnet}$& - &  NA & NA & $65,725,604$ & $84.02$  & $577,977,384$  & $19.98$ & -  \\
\hspace{0.1cm} N$3$HL$_\mathrm{gpt}$& $11,929,930$ & $\textbf{96.18}$ & $\textbf{91.65}$ & $24,950,436$ & $\textbf{85.68}$  & $113,547,048$  & $31.78$ & -  \\
\\
\hline

\multicolumn{4}{l}{\emph{   }}\\
\hline
\hspace{0.1cm} ResNet152 \textbf{(base 3)} & $58,164,298$ & $ 96.33$ & $90.89$ & $58,348,708$ & $86.24$ & $60,192,808$ & $31.12$ & $79.05$ \\
\hspace{0.1cm} ResNet200 & - & - & - & $62,829,732$ & $87.02$ & $64,673,832$  & $31.97$ & $\textbf{79.55}$\\
\hspace{0.1cm} N$3$HL152$_\mathrm{gpt}$ & -  & - & -  & $59,586,212$ & $\textbf{87.08}$ & $148,182,824$ & - & $78.50$ \\
\cmidrule(lr){1-1} \cmidrule(lr){2-4} \cmidrule(lr){5-6} \cmidrule(lr){7-9}
\multicolumn{1}{c}{} & \multicolumn{3}{c}{\thead{10 categories}} & \multicolumn{2}{c}{\thead{100 categories}} & \multicolumn{3}{c}{\thead{1000 categories}}
\end{tabular}
}
\caption{The image classification results are divided into three table sections, each compared with its own respective baseline for fair evaluation. Top (base 1): External signals on PHL$_\mathrm{small}$ simulate top-down attention. Middle (base 2): NHL$_\mathrm{gpt}$ vs. ResNet50. Bottom (base 3): NHL152$_\mathrm{gpt}$vs. ResNet152. Symbol - is used to exclude irrelevant data.}
\label{tab:tab}
\end{table*}

\begin{figure*}[b!]
    \centering
    \includegraphics[width=1\textwidth]{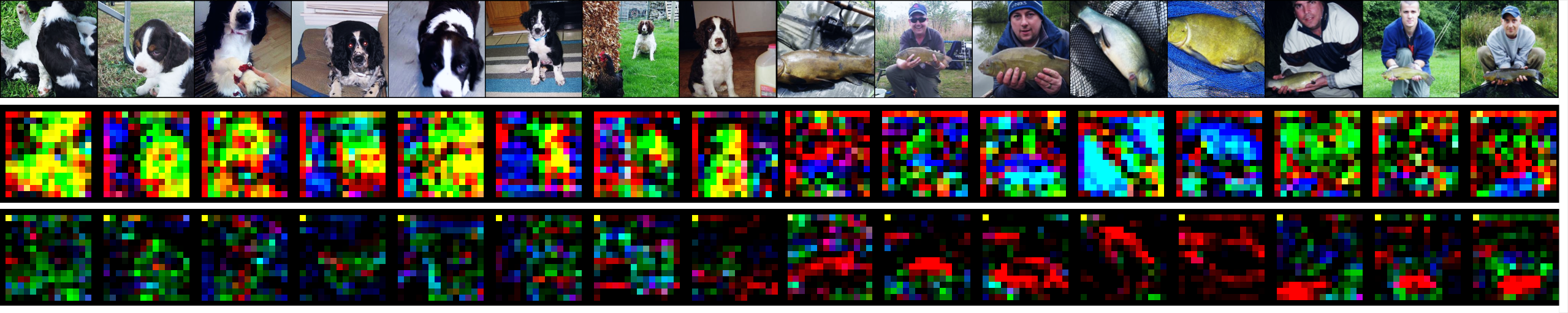}
    \caption{Category-specific features and original images from Imagenette \cite{imagenette}. The top row displays the images at a reduced resolution. The second and third rows show $14 \times 14 \times3$ RGB-colored, category-specific high-level features extracted using PHL$_\mathrm{big}$ - conv 4. Specifically, the features in the second row are from the ``dog'' branch, while the features in the third row from the ``fish'' branch. The selected feature maps from the dog branch are salient for the dog object and less salient for the fish object. The opposite is observed for the fish branch.}
\label{fig:saliency}
\end{figure*}

The cutouts presented in Sec.~\ref{ssec:NetArchi} are a unique property of PHL or NHL, where parameters are reduced by focusing only on the remaining and relevant categories, adapting to a changing task without requiring re-training. 
For the evaluation in ~\autoref{tab:cutouts}, we have divided the datasets into two subsets: lower label categories (e.g., category-ID: $1$-$5$) and upper label categories (e.g., category-ID: $6$-$10$). 
To start somewhere, we divide the 10 categories $50$:$50$ and the 100 categories $80$:$20$.
For the baseline, we have ResNets that are trained like the cutouts on the full dataset and validated only on the subset, and ResNets that are completely new trained only on the subset.
For Imagenette upper and lower $5$ the cutouts have a better top-1 acc. than newly trained ResNets
Slightly better are the ResNets, also trained on 10 categories, but the cutouts are able to reduce the parameters by approximately $28\,\%$.
To determine the computational complexity, we count giga-multiply-accumulate operations (GMACs) and find that we can save about $8 \, \%$ while maintaining very high performance.
In Imagewoof, the cutouts have a top-1 acc. with high values for the upper and a little worse for the lower 5, but when considering both accuracies its better and again with a reduction of parameters and GMACs. 
For ImageNet100 we show a cutout of $20$ categories. It's the same cutout as used in the visualization in \autoref{fig:cutout}. The cutout has the best result only slighlty increasing the GMACs but with a reduction of about $10$ million parameters. The last cutout at the bottom shows an evaluation from 
ImageNet1k. 

It is a good result for our method in terms of accuracy, due to the same and a little better acc. than the baseline, but with fewer GMACs, fewer parameters and less depth.
More importantly, it builds on the example of cross-modal interaction presented in the introduction of our paper by Marian et al.
\cite{crossmodalinteraction}, our approach can use already interpreted semantic cues such as the sound ``meow'' to focus visually. This allows the model to efficiently narrow down from $1000$ learned categories of ImageNet to the $5$ relevant ones, representing domestic cats. This demonstrates the practical application of our method with a high dynamic range of inference costs.
\subsection{Image Classification}
\label{ssec:imageclass}
For this task there are three baselines, so for better comprehensibility the \autoref{tab:tab} is also split into three sections:

\textbf{Base 1:} For $10$ categories the PHL nets have a top-1 acc. below the ResNets. 
This can be seen in the top section. When scaling the PHL to $100$ or $1000$ categories, the performance drops even more. 

\textbf{Base 2:} The later split-point and nesting show significant progress and lift our approaches N3HL und N4HL$_\mathrm{gpt}$ above the ResNet50 for $10$ and $100$ categories, only for $1000$ the methodology remains below. 
However, this may be due to the fact that Imagenet mini has so few training and validation images, which probably has a negative effect on our approach due to the higher number of parameters for $1000$ categories.
N2HL$_\mathrm{wordnet}$ cannot keep up, because the split-point was set too early or the nesting is not so advantageous. 

\textbf{Base 3:} Our N3HL153$_\mathrm{gpt}$ method can also outperform deeper networks such as the ResNet152 and the even deeper ResNet200 for Imagenet100. To show that the N3HL152$_\mathrm{gpt}$ remains competetive even with $1000$ categories, we trained the methods on ImageNet1k and ours is ahead of the similarly deep ResNet152.

\begin{table*}[t]
\centering
\resizebox{2.0\columnwidth}{!}{
\begin{tabular}{|l| ccc| cc| cc|}
\multicolumn{1}{c}{\thead{Dataset}} & \multicolumn{1}{c}{\thead{}} & \multicolumn{1}{c}{\thead{Imagenette}} & \multicolumn{1}{c}{\thead{Imagewoof}} & \multicolumn{1}{c}{} & \multicolumn{1}{c}{\thead{ImageNet100}} & \multicolumn{1}{c}{} & \multicolumn{1}{c}{\thead{ImageNet mini}}  \\
\cmidrule(lr){1-1} \cmidrule(lr){2-4} \cmidrule(lr){5-6} \cmidrule(lr){7-8}

\thead{Metric} & no. parameter & top-1 acc. & top-1 acc. & no. parameter & top-1 acc. & no. parameter & top-1 acc. \\
\hline
\multicolumn{4}{l}{\emph{Study 1}}\\
\hline
\hspace{0.2cm} PHL$_\mathrm{big}$ & $27,464,778$ &              $\textbf{95.75}$ & $\textbf{90.02}$ & $261,643,428$ &         $82.26$ & $2,603,429,928$  & {***} \\
\hspace{0.2cm} PHL$_\mathrm{small}$ & $9,611,338$ &               $95.71$ & $89.97$ & $83,109,028$ &           $\textbf{82.52}$ & $818,213,928$  & $21.10$ \\
\hline
\multicolumn{4}{l}{\emph{Study 2}}\\
\hline
\hspace{0.2cm} PHL$_\mathrm{lateSP}$ & $19,986,506$ & $\textbf{96.28}$ & $\textbf{90.86}$ & $122,975,396$ & $\textbf{84.49}$  & $1,152,864,296$  & $20.57$ \\
\hspace{0.2cm} PHL$_\mathrm{1branch}$& $9,688,778$ & {$95.90$} & {$90.61$} & $9,700,388$ & $83.94$  & $9,816,488$  & $\textbf{30.49}$\\
\hline
\multicolumn{7}{l}{\emph{Study 3}} \\
\hline
\hspace{0.2cm}NHL$_\mathrm{\overline{gpt}}$ & $11,929,930$ & $\textbf{96.25}$ & $91.47$ & $24,950,436$ & $85,50$ & $113,547,048$ &  {$30.48$} \\ 
\hspace{0.2cm}NHL$_\mathrm{gpt}$& $11,929,930$ & $96.18$ & $\textbf{91.65}$ & $24,950,436$ & $\textbf{85.68}$  & $113,547,048$   & {$\textbf{31.78}$}\\ 
\hline
\multicolumn{1}{l}{\emph{Study 4}} & \multicolumn{1}{c}{\thead{}} & \multicolumn{1}{c}{\thead{}} & \multicolumn{1}{c}{\thead{}} & \multicolumn{1}{c}{} & \multicolumn{1}{c}{\thead{}} & \multicolumn{1}{c}{} & \multicolumn{1}{c}{\thead{ImageNet1K}}  \\
\hline
\hspace{0.2cm}ConvNeXtV2-tiny$_\mathrm{base}$ & $27,874,186$ & $99.71$ & $\textbf{96.23}$ & $27,943,396$ & $90.53$ & $28,635,496$ & $81.07$\\
\hspace{0.2cm}ConvNeXtV2-tiny$_\mathrm{NHL}$& $15,277,834$ & $\textbf{99.72}$ & $96.22$ & $27,744,740$ & $\textbf{90.72}$  & $121,381,352$  & \textbf{81.10}\\
\cmidrule(lr){1-1} \cmidrule(lr){2-4} \cmidrule(lr){5-6} \cmidrule(lr){7-8}
\multicolumn{1}{c}{} & \multicolumn{3}{c}{\thead{10 categories}} & \multicolumn{2}{c}{\thead{100 categories}} & \multicolumn{2}{c}{\thead{1000 categories}}\\
\end{tabular}
}
\caption{Ablation Studies: This table presents the results of four focused studies. Study 1 examines the size of the architecture (big vs. small) in terms of parameter effectiveness. Study 2 evaluates the benefits of a later split-point and the parallel structure over a single branch sequential approach. Study 3 evaluates the effectiveness of nested features for ``similar'' categories versus nested ``non-similar'' categories. Study 4 tests our approach on a pre-trained model.
}
\label{tab:tababla}
\end{table*}

\subsection{Built-in Top-Down Attention} 
\label{ssec:selattinbuilt}
Consider a task where objects shall be classified in images and other modalities such as audio or descriptions are already available and provide prior semantic knowledge. This leads to contextual cues that can positively influence the classification, e.g. a ``meow'' should strengthen the classification of the object ``cat'' as in human perception \cite{Wolfe2021}.

We argue that the proposed PHL has a built-in selective attention mechanism because we know where to access the category-specific high-level features, since individual feature maps can be associated with a branch, its parameters, and a category.
~\autoref{fig:saliency} visualizes the feature maps of two parallel branches and shows category-specific differences.
In an initial test, we assume that meaningful semantic knowledge already exists for each validation image, so that amplification should be done for each image category.
This is done by simply multiplying the feature maps in the corresponding path in conv block 5 of PHL$_\mathrm{small}$ (see~\autoref{tab:bigtab}) by a scalar. Here we use values of $1.8$ (Imagenette), $1.4$ (Imagewoof), $1.1$ (ImageNet100) and $1.3$ (ImageNet mini), which is determined by experimentation, as there is a value at which the result can be influenced to the maximum; as soon as this value is exceeded, the influence ebbs away again. 
For the evaluation, the labels are used in such a way that a guidance exists that allows to find only the category-specific path of the respective validation image and to multiply the category-specific feature map from the conv layer 5b with the scalar (c.f.~\ref{tab:bigtab}).
This is done for all validation images and
the first section in \autoref{tab:tab} shows that the classification result of PHL$_\mathrm{small}$ is positively influenced. It is worth noting that this can be applied to any PHL or NHL network, like in \autoref{tab:cutouts}, highlighted by the light gray color.
This novel built-in top-down attention should be further explored, as it is not available in conventional architectures without much more complex operations \cite{kuo2022inferring}.

\subsection{Ablations}
\label{ssec:ablations}
Several ablation studies were required to make our methods competitive. The three most important are listed in \autoref{tab:tababla}. Study 1 was done to find an appropriate parameter setting for the parallel branches. Since PHL$_\mathrm{small}$ performs as well as PHL$_\mathrm{big}$, despite significantly fewer parameters, this suggests that PHL$_\mathrm{big}$ is oversized.
Study 2 shows an advantage of the SeqPar structure over the conventional sequential one, by PHL$_\mathrm{lateSP}$ achieving a higher acc. over the single-branch PHL$_\mathrm{1branch}$ for all datasets except ImageNet mini. We assume that too many parameters are detrimental for ImageNet mini, especially due to the small number of images per category. This makes the stability of the PHL remarkable for parameters up to the billion range. It is worth mentioning that for a 10-categories setting, PHL$_\mathrm{lateSP}$ outperforms a comparable ResNet50 with fewer parameters, indicating a good split-point location. Study 3 evaluates the effectiveness of nested features for ``similar'' categories versus nested ``non-similar''
categories. 
The differences are not as significant as we expected, but there is still an improvement in the nesting of similar features. The last shown study includes testing out topologies on a ConvNeXtV2 \cite{Woo2023ConvNeXtV2} architecture with frozen pretrained weights in the non-parallel part of the model. This simultaneously shows that our topologies can easily be implemented on more modern network architectures and that pretrained weights can be used effectively, only replacing part of the model with a parallel classification head without the need for retraining the rest of the model.
\section{Conclusion}
In this work, we have introduced a novel network topology that seamlessly integrates dynamic inference cost with a top-down attention mechanism. Inspired by the perceptual capabilities of the human brain, we combine a conventional sequential structure for the low- and mid-level features with parallelism and nesting for higher-level features based on ResNet50 for exemplary testing. Since this is a basic principle, it should be possible to apply it to the latest, already effective and/or unsupervised methods.

In terms of dynamic inference cost our methodology can achieve an exclusion of up to $73.48\,\%$ of parameters and $84.41\,\%$ fewer GMACs, analysis against comparative baselines show an average reduction of $40\,\%$ in parameters and $8\,\%$ in GMACs across the cases we evaluated.
These experiments indicate that the SeqPar structure, paves the way for AI models that are not only competetive, but also more interpretable, energy efficient, and adaptive. 

This advancement holds substantial promise for mobile computing, industrial, drone, robotic, and edge device applications, where computational resources are often limited.
Future research could further refine these topologies, potentially leading to breakthroughs in artificial intelligence that more closely resemble human cognitive processes.

\vspace{12pt}
\noindent
\emph{Acknowledgement:} Funded by the Deutsche Forschungsgemeinschaft (DFG, German Research Foundation) in the project Crossmodal Learning, TRR 169.

{\small
\bibliographystyle{ieee_fullname}
\bibliography{egbib}
}
\end{document}